\documentclass[pdflatex]{sn-jnl}% Style for submissions to Nature Portfolio journals
\usepackage{graphicx}%
\usepackage{multirow}%
\usepackage{amsmath,amssymb,amsfonts}%
\usepackage{amsthm}%
\usepackage{mathrsfs}%
\usepackage[title]{appendix}%
\usepackage{xcolor}%
\usepackage{textcomp}%
\usepackage{manyfoot}%
\usepackage{booktabs}%
\usepackage{algorithm}%
\usepackage{algorithmicx}%
\usepackage{algpseudocode}%
\usepackage{listings}%
\usepackage{lineno}%
\usepackage[most]{tcolorbox}
\usepackage{changepage}
\usepackage{soul}
\usepackage{scalerel}

\newcommand{\myemoji}[1]{\scalerel*{\includegraphics{#1.png}}{X}}

\theoremstyle{thmstyleone}%
\theoremstyle{thmstyletwo}%

\theoremstyle{thmstylethree}%

\begin{document}

\title[Article Title]{Design and evaluation of an agentic workflow for crisis-related synthetic tweet datasets}


\author*[1,3]{\fnm{Roben} \sur{Delos Reyes}}\email{rdelosreyes@student.unimelb.edu.au}

\author[2,3]{\fnm{Timothy} \sur{Douglas}}

\author[3]{\fnm{Asanobu} \sur{Kitamoto}}

\affil[1]{\orgdiv{School of Computing and Information Systems}, \orgname{The University of Melbourne}, \orgaddress{\street{Parkville}, \state{Victoria}, \country{Australia}}}

\affil[2]{\orgdiv{Department of Computer Science}, \orgname{University College London}, \orgaddress{\city{London}, \country{United Kingdom}}}

\affil[3]{\orgname{National Institute of Informatics}, \orgaddress{\city{Tokyo}, \country{Japan}}}


\abstract{Twitter (now X) has become an important source of social media data for situational awareness during crises. Crisis informatics research has widely used tweets from Twitter to develop and evaluate artificial intelligence (AI) systems for various crisis-relevant tasks, such as extracting locations and estimating damage levels from tweets to support damage assessment. However, recent changes in Twitter's data access policies have made it increasingly difficult to curate real-world tweet datasets related to crises. Moreover, existing curated tweet datasets are limited to past crisis events in specific contexts and are costly to annotate at scale. These limitations constrain the development and evaluation of AI systems used in crisis informatics. To address these limitations, we introduce an agentic workflow for generating crisis-related synthetic tweet datasets. The workflow iteratively generates synthetic tweets conditioned on prespecified target characteristics, evaluates them using predefined compliance checks, and incorporates structured feedback to refine them in subsequent iterations. As a case study, we apply the workflow to generate synthetic tweet datasets relevant to post-earthquake damage assessment. We show that the workflow can generate synthetic tweets that capture their target labels for location and damage level. We further demonstrate that the resulting synthetic tweet datasets can be used to evaluate AI systems on damage assessment tasks like geolocalization and damage level prediction. Our results indicate that the workflow offers a flexible and scalable alternative to real-world tweet data curation, enabling the systematic generation of synthetic social media data across diverse crisis events, societal contexts, and crisis informatics applications.} 


\maketitle

\section{Introduction}\label{intro}
Social media have become important sources of situational information in preparing for, responding to, and recovering from crises \cite{alexander2014social, houston2015social, olteanu2015expect, saroj2020use, erokhin2024social}. During such events, social media users share first- and second-hand observations about the crisis and its impacts, including news and announcements, reports of infrastructure damage and community needs, and personal experiences and sentiments. Crisis informatics research has widely utilized these data to enhance situational awareness before, during, and after crisis events \cite{palen2008online, reuter2018fifteen}. Leveraging advances in artificial intelligence (AI), many AI systems have been developed to automate the extraction of crisis-relevant information from social media data, supporting tasks such as early warning, event monitoring, behavioural analysis, misinformation detection, damage assessment, disaster report generation, and question answering \cite{imran2015processing, li2023exploring, ma2024surveying, xu2025large, lei2025harnessing}. 

Among social media data, tweets from Twitter (now X) have become a primary data source for crisis informatics studies \cite{ma2024surveying}. Twitter previously enabled broad access to historical and real-time tweets, allowing researchers to curate tweet datasets from crisis events and develop and experiment with various AI systems \cite{imran2015processing, li2023exploring, ma2024surveying, lei2025harnessing, alam2021crisisbench}. However, recent changes to Twitter's data access policies have made tweet collection increasingly restricted and costly \cite{bruns2021after, murtfeldt2024rip}. Moreover, existing curated tweet datasets are tied to specific historical crisis events and societal contexts, whose characteristics may not be applicable to future crises or different settings \cite{alam2021crisisbench}. In addition, annotating these datasets for developing and evaluating AI systems requires substantial manual effort, making them expensive to produce and difficult to scale \cite{hu2024dlrgeotweet}. As a result, many existing studies rely on relatively small labeled datasets that were curated for specific contexts and applications \cite{li2023exploring, lei2025harnessing, alam2021crisisbench, hu2024dlrgeotweet}. 

To address these limitations, synthetic data generation approaches have been used to produce synthetic data that can complement real-world data \cite{lu2023machine}. In particular, large language models (LLMs) have demonstrated the ability to generate realistic synthetic text across a wide range of applications, such as the natural sciences, software development, and healthcare \cite{long2024llms, naduas2025synthetic, alismail2025survey}. Their generative capabilities enable the scalable creation of data that can be customized to reflect specific characteristics tailored to particular purposes. For example, recent works used LLMs to generate synthetic tweets for experimenting with AI systems for sentiment classification and building function classification \cite{chim2025evaluating, bai2025generating}. LLM-driven generation thus creates opportunities to construct datasets in crisis contexts where real-world data are scarce, inaccessible, or unavailable.

However, LLMs as generative models have well-recognized limitations. Although capable of producing fluent text, LLMs are susceptible to errors and biases, requiring careful validation of their outputs \cite{ji2023survey, bender2021dangers}. Hence, while LLMs have been used for synthetic text generation, it is unclear whether they can reliably generate crisis-related tweets that must reflect the situational context and impact of a crisis event. In addition, LLMs alone do not inherently ensure that their outputs are correct or free from bias. Agentic design has emerged as an approach to address this challenge, in which an LLM is embedded within a workflow that guides its generation process. In such agentic workflows, multiple agents with explicit roles coordinate to perform the task, often incorporating feedback mechanisms that enable the automated evaluation and refinement of LLM-generated outputs \cite{alismail2025survey}.

In this study, we address challenges associated with curating real-world tweet datasets for crisis informatics research by developing a flexible and scalable agentic workflow for generating crisis-related synthetic tweet datasets. In particular, we make the following contributions:
\begin{itemize}
    \item We propose a novel agentic workflow that enables the iterative generation of crisis-related synthetic tweets. The workflow includes three interacting agents: a generator that uses an LLM to produce synthetic tweets conditioned on prespecified target labels, an evaluator that assesses these synthetic tweets using predefined heuristic-based compliance checks, and an augmenter that provides structured feedback from these evaluations to the generator to guide subsequent iterations.
    \item We empirically demonstrate that the proposed workflow can consistently generate synthetic tweets with controllable labels for location and damage level across multiple crisis events, through a post-earthquake damage assessment case study involving six real-world earthquake events with diverse contexts and impacts. 
    \item We show that the resulting synthetic tweet datasets capture key characteristics of real-world tweet datasets and can serve as a proxy for evaluating AI systems in geolocalization and damage level prediction tasks, highlighting the potential of the proposed workflow as a practical approach for generating synthetic social media data to support crisis informatics studies when real-world data are limited or unavailable.
\end{itemize}

The remainder of this paper is structured as follows. Section \ref{sec:methods} details the agentic workflow for generating crisis-related synthetic tweet datasets. Section \ref{sec:results} presents the results of the experiments in which the workflow is applied to generate synthetic tweet datasets for post-earthquake damage assessment. Section \ref{sec:discussion} discusses the implications and limitations of the proposed workflow. Finally, Section \ref{sec:conclusions} concludes with a summary of the study. 

\section{Methods}\label{sec:methods}
In this section, we present the proposed workflow for generating crisis-related synthetic tweet datasets. We begin by providing an overview of the workflow's agentic design. We then discuss the application of the workflow for a post-earthquake damage assessment case study. 

\subsection{Agentic workflow overview}   
We design an agentic workflow with the aim of generating a synthetic tweet dataset $\mathcal{D}_{syn}$ that can support crisis informatics research in contexts where real tweet datasets are inadequate or unavailable. The agentic workflow includes three agents: (1) a generator $g$, (2) an evaluator $e$, and (3) an augmenter $a$. The generator $g$ is responsible for generating a synthetic tweet $s$ conditioned on prespecified target labels. The evaluator $e$ then evaluates the synthetic tweet $s$ using predefined compliance checks. If the synthetic tweet $s$ passes all checks, it gets accepted to the synthetic tweet dataset $\mathcal{D}_{syn}$. Otherwise, the augmenter $a$ produces a structured feedback that is provided to the generator $g$ as additional context for why the synthetic tweet $s$ was rejected by the evaluator $e$. The workflow is iterative and can run for $n$ rounds or once the synthetic tweet dataset $\mathcal{D}_{syn}$ satisfies the desired criteria. Each agent in the workflow can be developed depending on the use case and requirements of the synthetic tweet dataset $\mathcal{D}_{syn}$, as well as the available resources of the workflow user.

\begin{figure}[t!]
    \centering
    \includegraphics[width=\linewidth]{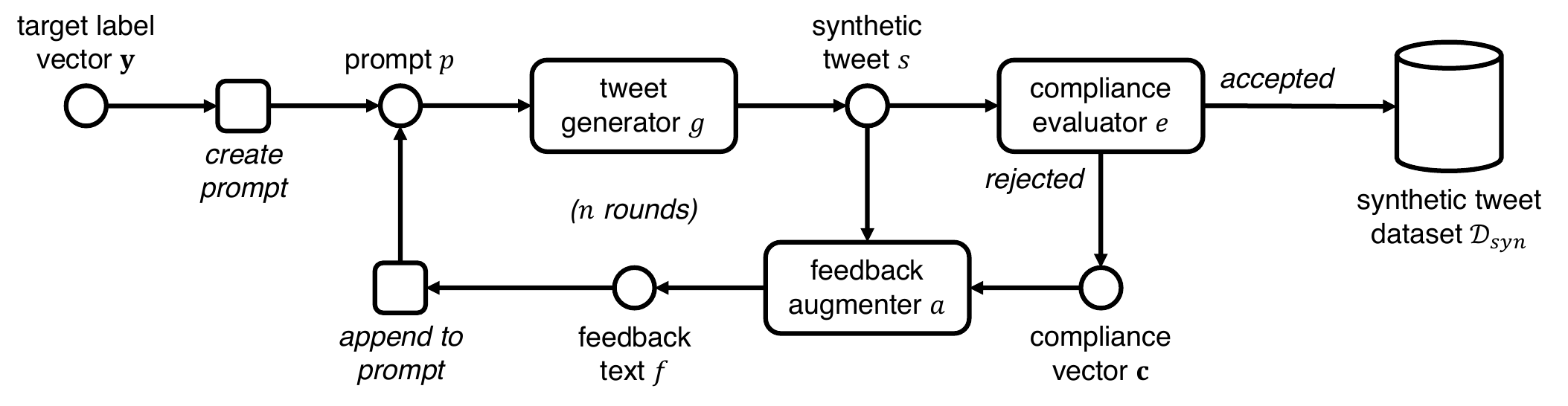}
    \caption{Schematic overview of the agentic workflow for generating crisis-related synthetic tweet datasets. The workflow includes three agents: (1) a tweet generator $g$ that generates synthetic tweets, (2) a compliance evaluator $e$ that evaluates those tweets, and (3) a feedback augmenter $a$ that provides feedback to the tweet generator $g$ based on the evaluations of the compliance evaluator $e$. The workflow iterates for $n$ rounds, with synthetic tweets that gets accepted by the compliance evaluator $e$ at each round added to the synthetic tweet dataset $\mathcal{D}_{syn}$.}
    \label{fig:fig1}
\end{figure}

Here, we apply the proposed workflow to rapidly yet robustly generate a synthetic tweet dataset relevant to post-earthquake damage assessment. In particular, we aim to generate synthetic tweets that reference a specific location and encapsulate a certain damage level, such that the resulting synthetic tweet dataset can be used for evaluating AI systems on two common damage assessment tasks. The first task is geolocalization, which has the objective of identifying the geographic locations referenced in tweets \cite{imran2015processing, ma2024surveying, xu2025large, lei2025harnessing, hu2024dlrgeotweet}. The second task is damage level prediction, where the aim is to classify the severity of damage described in tweets \cite{li2023exploring, ma2024surveying, xu2025large, lei2025harnessing, li2021social}. Ensuring that AI systems are accurate and reliable on these tasks across varying earthquake events is critical for supporting timely rescue operations and efficient resource allocations following earthquakes. To achieve this, we utilised an LLM as the generator $g$ to produce synthetic tweets with specific target labels for location and damage level (described in Section \ref{subsec:generator}). We then designed the evaluator $e$ to assess whether the synthetic tweets satisfy heuristic-based compliance checks (described in Section \ref{subsec:evaluator}). Finally, we developed the augmenter $a$ to output structured compliance messages for synthetic tweets rejected by the evaluator $e$, which gets appended as a feedback text to the prompt for the generator $g$ to guide subsequent rounds of the workflow (described in Section \ref{subsec:feedback}). We provide a schematic overview of this agentic workflow in Figure \ref{fig:fig1}.  

\subsection{Tweet generator}\label{subsec:generator}
The first agent in the workflow is the generator $g$, whose role is to generate synthetic tweets that capture their prespecified target labels. Here, we used an LLM for this agent, given the capability of LLMs to generate realistic texts and follow instructions \cite{long2024llms, chim2025evaluating, bai2025generating}. We express the target labels for each synthetic tweet using the target label vector $\textbf{y}$:
\begin{equation}\label{eq:target_label_vector}
    \textbf{y} = \{y_{loc},y_{dmg}\},
\end{equation}
where $y_{loc}$ is the target location and $y_{dmg}$ is the target damage level, which are labels relevant to the damage assessment tasks considered in this study. This target label vector $\textbf{y}$ is then encapsulated into the prompt $p$ that is provided to the LLM:
\begin{equation}
    p = q(\textbf{y}),
\end{equation}
where $q$ is a function for creating the prompt. Formally, we define the tweet generator $g$ to take as input a prompt $p$, embedded with the target labels from a target label vector $\textbf{y}$, and output a synthetic tweet $s$:
\begin{equation}
    s = g(p).
\end{equation}
As per the prompt template provided in Supplementary Section \ref{supp:prompt}, the LLM used as the tweet generator $g$ is instructed to generate a synthetic tweet, and the synthetic tweet $s$ is extracted from this LLM's output.

\subsection{Compliance evaluator}\label{subsec:evaluator}
The second agent in the workflow is the evaluator $e$, which is responsible for evaluating whether the synthetic tweets generated by the tweet generator $g$ satisfy predefined compliance checks. To enable fast turnaround times, we designed the evaluator to use quick yet robust heuristics. Here, we considered three heuristic-based compliance checks to assess whether each synthetic tweet $s$ satisfies the following: (1) location correctness, (2) damage level correctness, and (3) textual diversity. The location and damage level correctness checks determine whether the synthetic tweet $s$ reflects its prespecified target labels. On the other hand, the textual diversity check assesses whether the synthetic tweet $s$ is not overly similar to synthetic tweets already in the synthetic tweet dataset $\mathcal{D}_{syn}$. Based on these compliance checks, the compliance evaluator $e$ returns a compliance vector $\textbf{c}$ for a given a synthetic tweet $s$:
\begin{equation}
    \textbf{c} = e(s),
\end{equation}
such that 
\begin{equation}
    \textbf{c} = \{c_{loc}, c_{dmg}, c_{div}\},
\end{equation}
where $c_{loc}$, $c_{dmg}$, and $c_{div}$ are boolean values indicating compliance with the checks for location correctness, damage level correctness, and textual diversity, respectively. Only tweets that pass all checks (i.e., when $\textbf{c}=\textbf{1}$) are considered to be in compliance, and thus accepted by the compliance evaluator $e$ and get added to the synthetic tweet dataset $\mathcal{D}_{syn}$. 

\subsection{Feedback augmenter}\label{subsec:feedback}
The third agent in the workflow is the augmenter $a$. This augmenter is designed to convert the compliance vector $\textbf{c}$, returned by the compliance evaluator $e$, into a feedback text that is provided as additional context to the tweet generator $g$ for the next round of the workflow. In formal terms, the feedback augmenter $a$ receives as input the synthetic tweet $s$ and the compliance vector $\textbf{c}$, then returns a feedback text $f$:
\begin{equation}
    f = a(s, \textbf{c}).
\end{equation}
The feedback text $f$ consists of the synthetic tweet $s$ and compliance messages for any failed compliance checks, as listed in Table \ref{tab:feedback_text}. This feedback text $f$ is then appended to the prompt $p$:
\begin{equation}\label{eq:append_to_prompt}
    p = p \oplus f,
\end{equation}
in order to provide the tweet generator $g$ with information on how to refine its output in subsequent rounds, which could be seen as a form of in-context learning \cite{long2024llms}. We note that prompt $p$ accumulates the synthetic tweets and their associated compliance messages from the preceding rounds, which gives the tweet generator $g$ more examples to learn from with more iterations of the workflow. We provide examples of feedback texts in Supplementary Section \ref{supp:feedback_samples}. 

\begin{table}[t!]
    \centering
    \resizebox{\textwidth}{!}{
    \begin{tabular}{@{}p{4.5cm}p{7.5cm}@{}}
        \toprule
        \textbf{Compliance result} & \textbf{Compliance message}\\
        \midrule
        Incorrect location ($c_{loc}=0$) & Location ``\{location\_text\}'' not found in tweet\\
        Incorrect damage level ($c_{dmg}=0$) & Damage mismatch: expected ``\{target\_damage\_level\}'',\newline predicted ``\{kNN\_damage\_level\}''\\
        Lack diversity ($c_{div}=0$) & Too similar to accepted corpus\newline (Self-BLEU=\{bleu\_score\} $>$ \{bleu\_threshold\})\\
        \bottomrule
    \end{tabular}}
    \caption{Compliance results and their corresponding messages, which are included in the feedback text $f$ provided by the feedback augmenter $a$.}
    \label{tab:feedback_text}
\end{table}

\subsection{Preparing the target label vectors}
The agentic workflow requires as input a target label vector $\textbf{y}$, as illustrated in Figure \ref{fig:fig1}. To generate a synthetic tweet dataset $\mathcal{D}_{syn}$, the workflow thus requires a set of these target label vectors that characterizes the target label distribution in the synthetic tweet dataset $\mathcal{D}_{syn}$. These target label vectors can be extracted or derived from existing real-world tweets from past crisis events, representing observed or counterfactual target label distributions \cite{li2023exploring, lei2025harnessing, alam2021crisisbench}. Another option is to generate these target label vectors using crisis simulators, which are conventionally used to simulate crises and their impacts \cite{schneider2006hazus}.

Here, we used real tweet datasets containing tweets from past earthquake events that were collected in a previous study \cite{li2023exploring}. The earthquake events include: (1) the 2014 Napa, California earthquake, (2) the 2014 Iquique, Chile earthquake, (3) the 2015 Nepal earthquake, (4) the 2019 Ridgecrest, California earthquake, (5) the 2021 Fukushima, Japan earthquake, and (6) the 2021 Haiti earthquake. These earthquake events occurred in different geographic regions and time periods and had varying levels of impact as listed in Table \ref{tab:earthquake_summary}. As such, they provide diverse situational contexts related to earthquakes. These tweets were retrieved using Twitter's Search API on the day of each earthquake event and extending several days afterwards. Hence, the collected tweets reflect real-time observations during these earthquake events, which include location references, reports of damage, public reactions, and other situational updates shared by Twitter users. We removed duplicates and retweets, as the original tweet already describes the observation.

\begin{table}[t!]
    \centering
    \resizebox{\textwidth}{!}{
    \begin{tabular}{@{}p{3.5cm}p{1.5cm}p{1.5cm}p{1.5cm}p{1.5cm}p{1.3cm}@{}}
    \toprule
    \textbf{Earthquake\newline event} & \textbf{ShakeMap\newline intensity} & \textbf{DYFI\newline response} & \textbf{Fatality\newline count} & \textbf{Economic\newline loss} & \textbf{Damage\newline index}\\
    \midrule
    2014 Napa, California & VIII & VIII & Green & Red & 1.95\\
    2014 Iquique, Chile & VIII & IX & Yellow & Yellow & 2.12\\
    2015 Nepal & IX& VIII & Red & Red & 2.89\\
    2019 Ridgecrest, California & IX & VIII & Green & Yellow & 1.34\\
    2021 Fukushima, Japan & VII & IX & Yellow & Orange & 1.47\\
    2021 Haiti & IX & IX & Red & Red & 2.95\\
    \bottomrule
    \end{tabular}}
    \caption{Characteristics of the six real-world earthquake events considered in the experiments, spanning different years and geographic locations. Impacts of the earthquake events are described using ShakeMap intensities, Did You Feel It (DYFI) survey responses, estimated fatalities and economic losses, and Twitter-derived damage indices based on a four-level damage scale, as summarised in \cite{li2023exploring}.}
    \label{tab:earthquake_summary}
\end{table} 

Since these real-world tweets were unlabeled, we had to extract the target labels from each tweet in the real tweet datasets. To do this, we automatically extracted the target location $y_{loc}$ from each tweet using spaCy's \texttt{en\_core\_web\_trf} model \cite{honnibal2020spacy}, which is a pretrained named entity recognition model, and the target damage level $y_{dmg}$ using Google's \texttt{gemma-3-27b-it} model \cite{gemmateam2025gemma3technicalreport}. We used these models due to their strong performance in named entity recognition and instruction-following tasks, respectively, but other models with similar capabilities could also be used to annotate the unlabeled data. We note that if the real tweet datasets already have labels, then this automated annotation process would no longer be necessary since those labels can be used directly instead. We excluded tweets for which a target location $y_{loc}$ and a target damage level $y_{dmg}$ could not be extracted. These location-damage level pairs from the remaining tweets served as the target label vectors that were provided as input to the agentic workflow.

\subsection{Experimental setup}
To evaluate the ability of the agentic workflow to generate synthetic tweet datasets for post-earthquake damage assessment and to assess the utility of these synthetic datasets for evaluating AI systems on geolocalization and damage level prediction tasks, we ran the workflow using the target label vectors obtained from each of the six earthquake events. For the agentic workflow, we implemented each agent as described below.

\subsubsection{Choosing the tweet generator}
For the tweet generator $g$, we used Google's \texttt{gemma-3-1b-it} model \cite{gemmateam2025gemma3technicalreport}. Based on our preliminary experiments, we found that this lighter \texttt{gemma-3-1b-it} model is sufficient for tweet generation similar to other heavier Gemma 3 models, while requiring less computation and enabling faster generation. We also tested with using other lightweight models such as Qwen's \texttt{Qwen-3-0.6B} model and Meta's \texttt{Llama-3.2-1B-Instruct} model as the tweet generator $g$, which also worked well like the \texttt{gemma-3-1b-it} model as discussed in Supplementary Section \ref{supp:llms} and can be used alternatively.

\subsubsection{Implementing the compliance checks}
We implemented the three heuristic-based compliance checks used by the compliance evaluator $e$ as follows:

\paragraph{Location correctness check}
We identified whether the target location $y_{loc}$ appeared in the synthetic tweet $s$ using a case-insensitive substring matching. If the lowercased target location $y_{loc}$ is present in the lowercased synthetic tweet $s$, then the synthetic tweet $s$ passes the location correctness check.

\paragraph{Damage level correctness check}
We first estimated the damage level of the synthetic tweet $s$ based on the cosine similarity between its tweet embedding and embeddings from the labeled, real tweet datasets, which serve as reference embeddings. We used the k-nearest neighbors (kNN) algorithm to assign the damage level of the synthetic tweet $s$ based on this embedding similarity using $k=5$ and embeddings provided by \texttt{all-mpnet-base-v2} \cite{reimers2019sentencebert}, a commonly used sentence transformer model. If this kNN-based damage level of the synthetic tweet $s$ is consistent with its target damage level $y_{dmg}$, then the synthetic tweet $s$ passes the damage level correctness check.

\paragraph{Textual diversity check}
We quantified the textual diversity of the synthetic tweet $s$ using self-BLEU scores, which measure the 4-gram similarity between tweets \cite{chim2025evaluating, bai2025generating}. This check ensures that the synthetic tweet dataset $\mathcal{D}_{syn}$ encompasses a wide range of textual descriptions that provide a reasonable level of linguistic complexity necessary when assessing damage from tweets. The synthetic tweet $s$ passes the textual diversity check if it has a self-BLEU score below a predetermined self-BLEU threshold $\ell$, noting that a self-BLEU score lower than this threshold indicates low similarity to synthetic tweets already in the current synthetic tweet dataset $\mathcal{D}_{syn}$. We set $\ell=40$ based on the self-BLEU scores observed in the real tweet datasets and computed the self-BLEU score for each synthetic tweet $s$ against 100 synthetic tweets randomly sampled from the current synthetic tweet dataset $\mathcal{D}_{syn}$.

\subsubsection{Appending feedback text to the prompt}
The feedback text $f$ returned by the feedback augmenter $a$ is added to the prompt $p$ provided to the tweet generator $g$ by extending the prompt with the following instruction:
\begin{tcolorbox}[breakable]
\setlength\linenumbersep{0.2cm}
\setcounter{linenumber}{1}
\begin{internallinenumbers}
The previous attempt was incorrect with the following feedback:\\
\{feedback text $f$\}\\

Regenerate the tweet correcting these issues.
\end{internallinenumbers}
\end{tcolorbox}
\noindent With each additional feedback round, the new feedback text $f$ is appended to this section of the prompt $p$ if the newly generated synthetic tweet $s$ fails one or more compliance checks. These accumulated feedback texts guide the tweet generator $g$ by providing examples of incorrect synthetic tweets and indicating which aspects should be corrected in the subsequent round.

\section{Results}\label{sec:results}
Here, we present the results of applying the agentic workflow to generate synthetic tweet datasets with target locations and damage levels from the six earthquake events. First, we evaluated the proposed workflow based on (1) how accurately it generated synthetic tweets that reflected their prespecified target labels and (2) how efficiently it produced these synthetic tweets. Second, we assessed the quality of the resulting synthetic tweet datasets in terms of (1) the similarity of their characteristics to those of the real tweet datasets and (2) their utility for evaluating AI systems in geolocalization and damage level prediction tasks.

\begin{figure}[t!]
    \centering
    \includegraphics[width=0.9\linewidth]{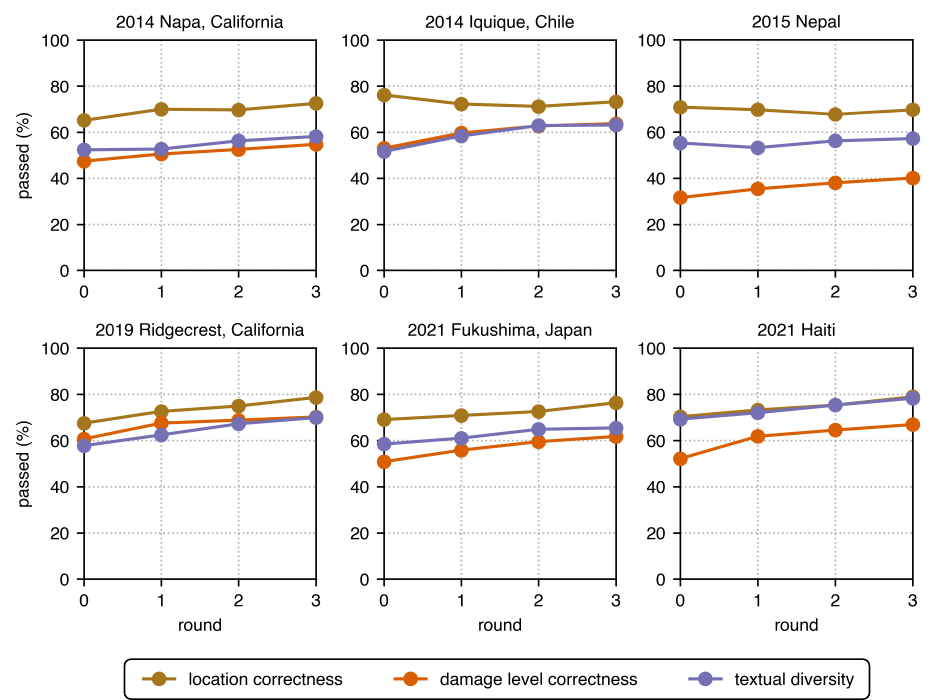}
    \vspace{0.4cm}
    \caption{Percentage of synthetic tweets that passed compliance checks for location correctness, damage level correctness, and textual diversity. In the initial generation attempt, at least 30\% of synthetic tweets passed each compliance check. This percentage improved with every additional feedback round.}
    \label{fig:fig2}
\end{figure}

\begin{figure}[t!]
    \centering
    \includegraphics[width=0.9\linewidth]{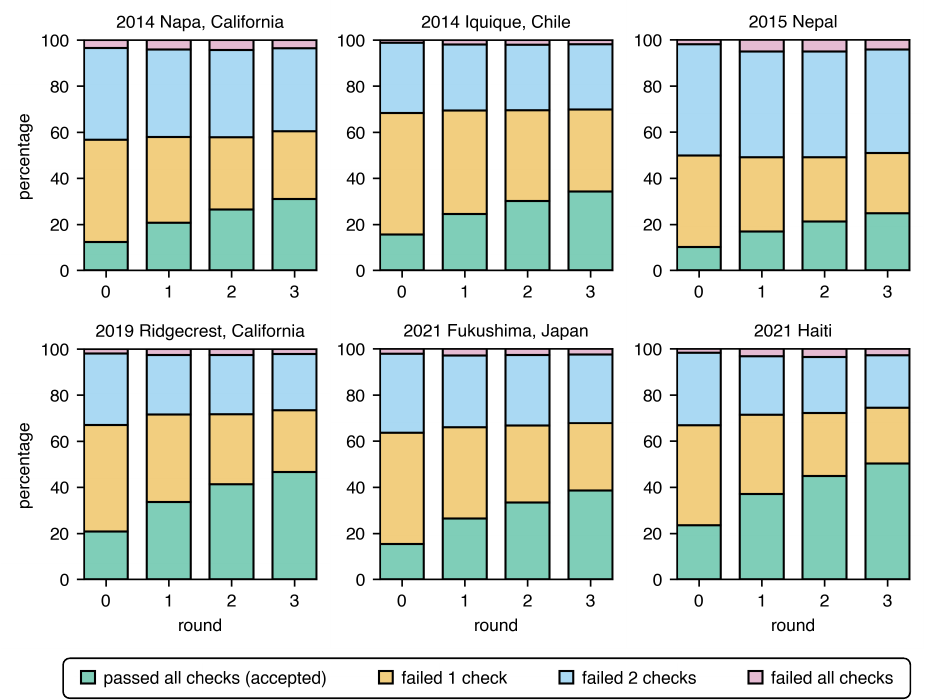}
    \vspace{0.4cm}
    \caption{Distribution of synthetic tweets by the number of compliance checks passed across the six earthquake events. The percentage of accepted synthetic tweets increases with more feedback rounds.}
    \label{fig:fig3}
\end{figure}

\subsection{Capturing the prespecified target labels}
To assess whether the workflow can generate synthetic tweets that captured their prespecified target labels, we measured the percentage of synthetic tweets that passed the compliance checks for location correctness and damage level correctness. Across all earthquake events, about 30-80\% of synthetic tweets passed each of these checks after the initial generation attempt as shown in Figure \ref{fig:fig2}, demonstrating the tweet generator's ability to generate tweets that reflect their target labels even without any feedback. Between these two correctness checks, the damage level correctness was more difficult to satisfy, having a lower passing rate in four of the six earthquake events. The workflow also generated synthetic tweets with correct locations and damage levels while maintaining textual diversity, as exhibited in Figure \ref{fig:fig2}.

Iterative feedback substantially improved both the correctness and diversity of the generated synthetic tweets. With each feedback round, the percentage of synthetic tweets passing each individual check increased in general across all events, as illustrated by the increasing trends in Figure \ref{fig:fig2}. Consequently, a growing percentage of synthetic tweets satisfied all three compliance checks with more feedback rounds, as exhibited in Figure \ref{fig:fig3}. After three feedback rounds, about 20-50\% of the synthetic tweets passed all three compliance checks and were added to the synthetic tweet datasets. Similar patterns were observed when alternative LLMs were used as the tweet generator as shown in Supplementary Figure \ref{fig:s_llms}, indicating that these improvements are not specific to the tweet generator but a result of the agentic and iterative design of the workflow. 

\begin{figure}[t!]
    \centering
    \includegraphics[width=0.68\linewidth]{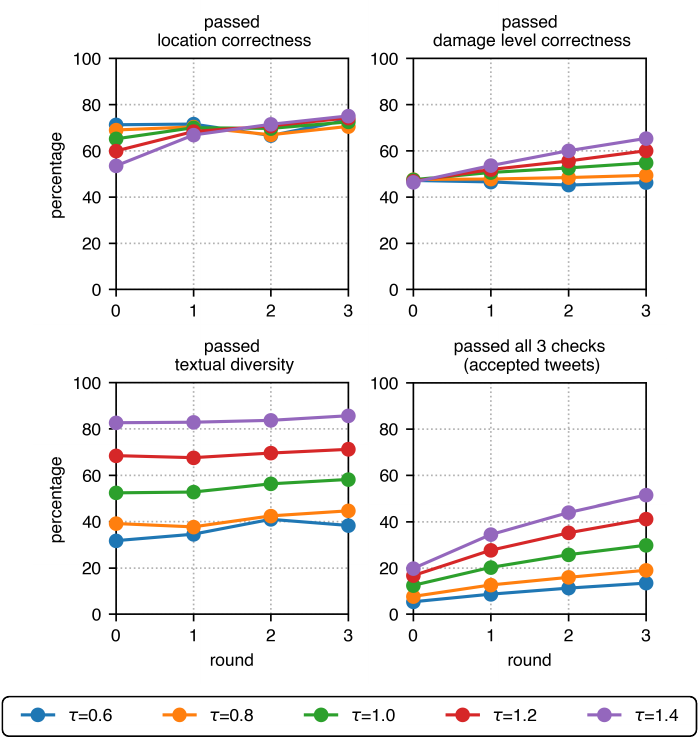}
    \vspace{0.4cm}
    \caption{Percentage of tweets that passed the compliance checks when using different temperature $\tau$ values. Higher temperature values generated more diverse synthetic tweets, resulting in more synthetic tweets that passed the textual diversity check and were eventually accepted and added to the synthetic tweet dataset. Results shown are based on the 2014 Napa, California earthquake event.}
    \label{fig:fig4}
\end{figure}

\subsection{Assessing workflow efficiency}
LLMs have a temperature $\tau$ parameter that controls the stochasticity of their outputs. In the experiments described above, the tweet generator used the default temperature $\tau=1.0$. To examine whether this parameter affects the efficiency of the workflow in terms of the quality and quantity of the synthetic tweet datasets, we ran an experiment in which we varied the temperature $\tau$ from 0.6 to 1.4, with increments of 0.2.

Higher temperature values resulted in a greater number of synthetic tweets being accepted at each feedback round and overall, as shown in Figure \ref{fig:fig4} for the 2014 Napa, California earthquake. Temperature had a small effect on improving location correctness and damage level correctness. In contrast, it had a large effect on textual diversity, where higher temperature values generated synthetic tweets that were more likely to satisfy the textual diversity check. While feedback rounds generally improved both the correctness and diversity of the synthetic tweets across temperature values, the temperature parameter had a stronger influence on textual diversity. Similar results were observed for the other earthquake events, as displayed in Supplementary Figure \ref{fig:s_temp}. 

\subsection{Analyzing synthetic dataset characteristics}
We analyzed the synthetic tweet datasets produced by the workflow with three feedback rounds ($n=3$) and a temperature $\tau=1.4$, which resulted in the largest number of accepted synthetic tweets for each of the six earthquake events. These synthetic tweet datasets exhibited distinct location and damage level characteristics as summarized in Table \ref{tab:syn_char}. The 2021 Haiti earthquake had the most number of unique locations (28\%), while the 2019 Ridgecrest, California earthquake had the least (9\%). Earthquake events with higher estimated fatalities and economic losses (i.e., the 2015 Nepal and 2021 Haiti earthquakes) were associated with a greater proportion of synthetic tweets that captured severe damage. In contrast, events with lower estimated losses (e.g., the 2019 Ridgecrest earthquake) were associated with more synthetic tweets representing lower damage levels. 

\begin{figure}[t!]
    \centering
    \includegraphics[width=\linewidth]{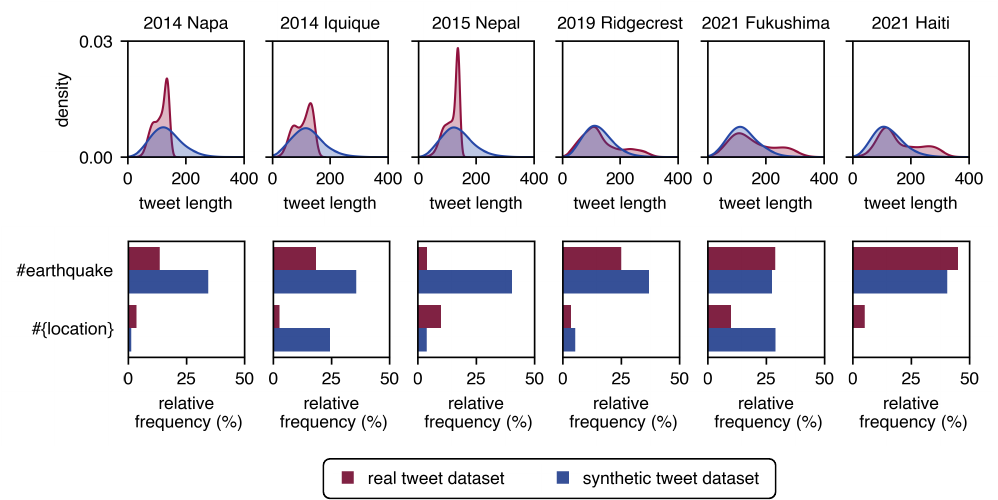}
    \caption{Tweet length distributions and relative hashtag frequencies in the real and synthetic tweet datasets. The synthetic tweets reflect common characteristics of real tweets, such as character limits and the use of common hashtags during crises (e.g., \#earthquake, \#napa, \#iquique, \#nepal, \#ridgecrest, \#fukushima, \#haiti).}
    \label{fig:fig5}
\end{figure}

To compare the structural characteristics of the synthetic tweet datasets with those of the real tweet datasets, we collected the real tweets from which the target labels for each accepted synthetic tweet were obtained. The distribution of tweet lengths was comparable between the real and synthetic tweet datasets across the six earthquake events, as shown in Figure \ref{fig:fig5}. Likewise, hashtag patterns (e.g., the use of \#earthquake during earthquake events) were reflected in the synthetic tweet datasets. Notably, both these characteristics were not explicitly specified during the generation process but were implicitly reproduced by the tweet generator.  

\begin{table}[t!]
    \centering
    \resizebox{\textwidth}{!}{
    \begin{tabular}{@{}lllllll@{}}
    \toprule
     & & \multicolumn{1}{l}{\textbf{Location}} & \multicolumn{4}{l}{\textbf{Damage level}}\\
    \cmidrule(lr){3-3} \cmidrule(lr){4-7} 
    \textbf{Event} & \textbf{Size} & \textbf{Unique} & \textbf{0} & \textbf{1} & \textbf{2} & \textbf{3}\\
    \midrule
    2014 Napa & 6,819 & 789 (12) & 3,841 (56) & 2,386 (35) & 592 (9) & 0 (0)\\
    2014 Iquique & 2,646 & 446 (17) & 2,238 (85) & 229 (9) & 179 (7) & 0 (0)\\
    2015 Nepal & 31,709 & 3,476 (11) & 9,581 (30) & 7,156 (23) & 12,114 (38) & 2,858 (9)\\
    2019 Ridgecrest & 58,495 & 5,325 (9) & 50,859 (87) & 6,600 (11) & 1,019 (2) & 17 ($<$1)\\
    2021 Fukushima & 2,547 & 340 (13) & 1,612 (63) & 807 (32) & 128 (5) & 0 (0)\\
    2021 Haiti & 8,105 & 2,230 (28) & 6,439 (79) & 225 (3) & 1,240 (15) & 201 (2)\\
    \bottomrule
    \end{tabular}}
    \caption{Characteristics of the synthetic tweet datasets, including dataset size, unique location counts, and frequency distribution of damage levels. The datasets comprise synthetic tweets produced by the workflow with three feedback rounds ($n=3$) and a temperature  $\tau=1.4$.}
    \label{tab:syn_char}
\end{table}

\begin{table}[t!]
    \centering
    \resizebox{\textwidth}{!}{
    \begin{tabular}{@{}p{1.8cm}p{1.4cm}p{1.4cm}p{1.4cm}p{1.4cm}p{1.4cm}p{1cm}@{}}
    \toprule
    \textbf{Earthquake\newline event} & 2014\newline Napa & 2014\newline Iquique & 2015\newline Nepal & 2019\newline Ridgecrest & 2021\newline Fukushima & 2021\newline Haiti\\
    \midrule
    \multicolumn{7}{@{}l}{\textbf{Geolocalization}}\\
    \midrule
    \multicolumn{7}{@{}l}{Percentage of geolocated tweets}\\
    \hspace{5pt} Real & 94.41 & 94.33 & 95.69 & 91.36 & 99.14 & 97.36\\
    \hspace{5pt} Synthetic & 95.76 & 72.56 & 89.83 & 85.71 & 93.95 & 85.44\\
    \multicolumn{7}{@{}l}{Accuracy relative to geolocated tweets}\\
    \hspace{5pt} Real & 64.59 & 32.85 & 73.63 & 76.48 & 75.37 & 74.21\\
    \hspace{5pt} Synthetic & 74.46 & 73.85 & 85.52 & 82.50 & 87.96 & 83.49\\
    \midrule
    \multicolumn{7}{@{}l}{\textbf{Damage level prediction}}\\
    \midrule
    \multicolumn{7}{@{}l}{Percentage of tweets with valid damage levels}\\
    \hspace{5pt} Real & 99.78 & 99.81 & 99.50 & 98.64 & 98.15 & 99.05\\
    \hspace{5pt} Synthetic & 98.75 & 98.83 & 98.85 & 98.22 & 97.96 & 98.47\\
    \multicolumn{7}{@{}l}{Accuracy relative to tweets with valid damage levels}\\
    \hspace{5pt} Real & 52.55 & 58.65 & 50.80 & 37.94 & 32.16 & 64.59\\
    \hspace{5pt} Synthetic & 38.00 & 28.60 & 56.56 & 28.70 & 35.39 & 36.70\\
    \bottomrule
    \end{tabular}}
    \caption{Evaluation of pretrained AI systems on geolocalization and damage level prediction tasks using the real and synthetic tweet datasets. The synthetic tweet datasets enabled the AI systems to identify locations and predict damage levels similarly to the real tweet datasets. AI system performance on the two tasks, measured by percentage of valid outputs and accuracy of these outputs relative to their target labels, was comparable when using the real and synthetic tweet datasets.}
    \label{tab:utility}
\end{table}

\subsection{Evaluating synthetic dataset utility} 
To examine whether these synthetic tweet datasets can be useful for experimenting with AI systems for crisis applications, we used them to evaluate the performance of pretrained AI systems on geolocalization and damage level prediction tasks. For the geolocalization task, we used spaCy's \texttt{en\_core\_web\_sm} model \cite{honnibal2020spacy}. For the damage level prediction tasks, we used Google's \texttt{gemma-3-4b-it} model \cite{gemmateam2025gemma3technicalreport}. We note that both these models are lighter versions of the \texttt{en\_core\_web\_trf} and \texttt{gemma-3-27b-it} models that were used to extract the location and damage level labels, respectively, improving computational efficiency while remaining suitable for evaluation purposes.

The synthetic tweet datasets enabled pretrained AI systems to identify locations and predict damage levels, as exhibited in Table \ref{tab:utility}. The geolocalization and damage level prediction results on the synthetic tweet datasets were also generally comparable to those obtained using the real tweet datasets, indicating that the synthetic tweet datasets could serve as a proxy for the real tweet datasets. The percentage of geolocated tweets identified by the \texttt{en\_core\_web\_sm} model across the six earthquake events ranged from 91.36\% to 99.14\% when using the real tweet datasets and ranged from 72.56\% to 95.76\% when using the synthetic tweet datasets. Geolocalization accuracy of the \texttt{en\_core\_web\_sm} model, relative to the geolocated tweets, ranged from 32.85\% to 76.48\% for the real tweet datasets and from 73.85\% to 87.96\% for the synthetic tweet datasets. Similarly, the percentage of tweets with valid damage levels predicted by the \texttt{gemma-3-4b-it} model ranged from 98.15\% to 99.81\% when using the real tweet datasets and ranged from 97.96\% to 98.85\% when using the synthetic tweet datasets. Damage level prediction accuracy of the \texttt{gemma-3-4b-it} model, relative to tweets with valid damage levels, ranged from 32.16\% to 64.59\% for the real tweet datasets and from 28.60\% to 56.56\% for the synthetic tweet datasets.

\section{Discussion}\label{sec:discussion}
Real-world social media data are key data sources in crisis informatics research and have been widely used in developing artificial intelligence (AI) systems designed to support decision making across diverse crisis contexts and applications \cite{imran2015processing, li2023exploring, ma2024surveying, xu2025large, lei2025harnessing}. In particular, tweets from Twitter (now X) have been a primary data source in existing crisis informatics studies and crisis-related social media datasets \cite{imran2015processing, ma2024surveying, alam2021crisisbench}. However, curating real-world tweet datasets is challenging because of limitations in data availability, collection, and annotation \cite{alam2021crisisbench, bruns2021after, murtfeldt2024rip, hu2024dlrgeotweet}. These real-world data limitations have made it difficult to develop and experiment with AI systems used in crisis informatics. To address these limitations, we developed an agentic workflow for generating crisis-related synthetic tweet datasets and demonstrated its application to a post-earthquake damage assessment case study. 

Our results demonstrate that the proposed workflow can effectively and efficiently generate synthetic tweet datasets relevant to its intended crisis use case. Given specific target labels for location and damage level, the workflow generated tweets that captured their prespecified target labels as shown in Figure \ref{fig:fig2}. The workflow thus enables the generation of synthetic tweets with controllable labels, mitigating the need to manually collect and annotate tweets to conduct crisis informatics studies. In addition, synthetic tweet datasets were adequately generated across six real-world earthquake events with differing characteristics, as listed in Table \ref{tab:earthquake_summary} and \ref{tab:syn_char}. While only about 50\% of the target labels resulted in accepted synthetic tweets, the synthetic tweet datasets did capture the distinct characteristics of each earthquake event. The number of rounds $n$ could, for example, be increased to expand the size of these synthetic tweet datasets if desired or be configured to iterate until the synthetic tweet datasets are deemed suitable for their intended use case. Furthermore, the synthetic tweet datasets could be used as a proxy for evaluating AI systems on two different damage assessment tasks, namely, geolocalization and damage level prediction, as exhibited in Table \ref{tab:utility}. This result indicates that our proposed workflow can support the development and evaluation of AI systems for crisis use cases where real-world data are insufficient or unavailable.

An important consideration when developing the workflow was the adoption of an agentic design \cite{alismail2025survey}. While generators like large language models (LLMs) are able to generate realistic synthetic text, they can generate incorrect or biased outputs \cite{ji2023survey, bender2021dangers}. The agentic design helps validate and refine the tweet generator's outputs over multiple rounds of the workflow, improving both the quantity and quality of the synthetic tweet datasets as shown in Figures \ref{fig:fig2} and \ref{fig:fig3}. This agentic design also makes the workflow modular, enabling it to be readily adapted or extended to meet specific conditions or requirements. For example, different LLMs can be substituted for the tweet generator, as shown in Supplementary Figure \ref{fig:s_llms}. Similarly, the temperature of the LLM-based tweet generator can be adjusted to improve the efficiency of the workflow as illustrated in Figure \ref{fig:fig4} and Supplementary Figure \ref{fig:s_temp}.

Although this study applied the agentic workflow to generate synthetic tweet datasets related to post-earthquake damage assessment, the proposed workflow can be readily adapted to other crisis events and crisis informatics applications. The overall design of the workflow, which uses a generator, evaluator, and augmenter in a feedback loop, is independent of the crisis context, but the design of each agent depends on the specific context. In particular, the target labels in the target label vector $\textbf{y}$, the prompt $p$, the compliance checks captured in the compliance vector $\textbf{c}$, and the corresponding compliance messages included in the feedback text $f$ should be modified when applying the workflow to other crisis events, such as floods and hurricanes, and crisis informatics applications, such as behavioural analysis or misinformation detection. Nonetheless, these modifications can be implemented straightforwardly given the workflow's agentic design.

Our study has several limitations. First, the current workflow requires target labels to guide the generation process, which assumes a certain level of prior knowledge about the crisis. However, such target labels can be obtained or derived from existing datasets from past crisis events or generated using crisis simulators \cite{li2023exploring, schneider2006hazus}. Second, the damage level correctness requires reference embeddings to evaluate the damage level of a synthetic tweet. Nonetheless, such reference embeddings can also be based on available labeled datasets \cite{lei2025harnessing, alam2021crisisbench}. Alternatively, a small, reliable, and general labeled dataset can be created to use as a reference for this compliance check. Finally, we implemented the generator, evaluator, and augmenter to prioritize the speed of the workflow for producing synthetic tweet datasets, which is desirable for systematic experimentation. However, this emphasis on speed may have reduced output quality, resulting in noisier synthetic tweet datasets. Several improvements could mitigate this issue, such as employing a more capable LLM as the tweet generator or using an LLM-based compliance evaluator to enable a more nuanced evaluation process, although these improvements would also require more computational resources.

\section{Conclusions}\label{sec:conclusions}
In summary, we introduced an agentic workflow for generating synthetic tweet datasets tailored to crisis informatics research, demonstrated through a case study on post-earthquake damage assessment. We showed that the workflow produces synthetic tweets that capture prespecified target characteristics. We further demonstrated that the resulting synthetic tweet datasets can be useful for evaluating AI systems on crisis-relevant tasks, such as geolocalization and damage level prediction. By enabling the controllable and scalable generation of social media data such as tweets, the workflow helps address limitations in curating real-world social media related to crises and supports the systematic generation of synthetic social media data across diverse crisis contexts and applications.

\backmatter

\section*{Funding}
Roben Delos Reyes is supported by the Melbourne Research Scholarship provided by the University of Melbourne. This study was conducted during the internships of Roben Delos Reyes and Timothy Douglas at the National Institute of Informatics (NII) and was supported by the NII International Internship Program.

\section*{Acknowledgements}
The experiments conducted in this study were supported by the resources provided by the University of Melbourne's Research Computing Services and the Petascale Campus Initiative. We would also like to thank Lingyao Li for sharing the real tweet datasets used in the experiments.

\section*{Data availability}
The synthetic tweet datasets produced in this study are available at: \url{https://github.com/rddelosreyes/synthetic-crisis-tweets}.

\bibliographystyle{unsrt}
\bibliography{sn-bibliography}

\newpage
\renewcommand{\thefigure}{S\arabic{figure}}
\renewcommand{\thetable}{S\arabic{table}}
\renewcommand{\thesection}{S\arabic{section}}
\setcounter{figure}{0}
\setcounter{table}{0}
\setcounter{section}{0}

\noindent{\LARGE Supplementary Material \par}

\section{Tweet generator prompt}\label{supp:prompt}

The prompt $p$ that is provided to the tweet generator $g$ encapsulates the target location $y_{loc}$ and the target damage level $y_{dmg}$ as follows:

\begin{tcolorbox}[breakable]
\setlength\linenumbersep{0.2cm}
\setcounter{linenumber}{1}
\begin{internallinenumbers}
Task:\\
You are a synthetic tweet generator. Your task is to generate a tweet as if it was posted by a real Twitter user after an earthquake event. Consider varying the persona that generated the tweet (eg, a concerned citizen expressing his/her sentiment, a government/news agency providing information, a user reporting his/her firsthand observations about the situation, etc.)\\

Instructions:\\
You must generate one synthetic tweet such that a large language model (LLM) can satisfy the following conditions:\\
1. Identify ``\{target location $y_{loc}$\}'' from the synthetic tweet\\
2. Identify that the tweet is related to damage from the earthquake event\\
3. Identify that the tweet has a damage level of ``\{target damage level $y_{dmg}$\}'' based on the following damage scale:
\begin{adjustwidth}{2em}{0em}
    0 - No damage or injury. This damage level corresponds to levels I-III in the Modified Mercalli Intensity (MMI) scale which has any of the following characteristics: no noticeable damage; felt by only a few people at rest; no damage to buildings; felt indoors, especially on upper floors; no significant structural damage.
\end{adjustwidth}
\begin{adjustwidth}{2em}{0em}
    1 - Slight damage. This damage level corresponds to levels IV-V in the Modified Mercalli Intensity (MMI) scale which has any of the following characteristics: felt by most people; some damage to buildings, such as minor cracks; felt by everyone; damage to buildings, minor cracks, but no collapse.
\end{adjustwidth}
\begin{adjustwidth}{2em}{0em}
    2 - Moderate damage with the possibility of injuries. This damage level corresponds to levels VI-VII in the Modified Mercalli Intensity (MMI) scale which has any of the following characteristics: damage to buildings, visible structural deformation; significant damage, some collapses or structural failures.
\end{adjustwidth}
\begin{adjustwidth}{2em}{0em}
    3 - Severe damage with the possibility of fatalities. This damage level corresponds to levels VIII-X in the Modified Mercalli Intensity (MMI) scale which has any of the following characteristics: many buildings collapse or are severely damaged; total destruction in some areas, severe damage; complete destruction of all structures in the affected area.\\
\end{adjustwidth}
\newpage
Input:\\
Location: \{target location $y_{loc}$\}\\
Damage level: \{target damage level $y_{dmg}$\}\\

Output:\\
Do not provide additional output except for the following, in strict JSON format:\\
\{\{
\begin{adjustwidth}{2em}{0em}
    ``synthetic\_tweet\_text'': ``$<$ synthetic tweet $>$''
\end{adjustwidth}
\}\}
\end{internallinenumbers}
\end{tcolorbox}

\section{Feedback text examples}\label{supp:feedback_samples}
Examples of the feedback text $f$ for rejected synthetic tweets, with varying compliance results across different rounds, are provided in Supplementary Table \ref{tab:feedback_examples}. As described in Equation \ref{eq:append_to_prompt}, the feedback text $f$ at each round gets appended to the prompt $p$. Hence, after each round, the prompt $p$ includes the feedback text $f$ listed in Supplementary Table \ref{tab:feedback_examples} for that round and the preceding rounds. This accumulation of feedback text, which is provided by the feedback augmenter $a$, gives the tweet generator $g$ more examples and context to guide its generation process every time it generates a synthetic tweet that gets rejected by the compliance evaluator $e$.

\begin{table}[h!]
    \centering
    \resizebox{\textwidth}{!}{
    \begin{tabular}{@{}p{0.8cm}p{0.8cm}p{0.8cm}p{0.8cm}p{8cm}@{}}
    \toprule
    \textbf{round} & $c_{loc}$ & $c_{dmg}$ & $c_{div}$ & \textbf{feedback text} $f$\\
    \midrule
    0 & 0 & 1 & 1 & Generated tweet: Just saw a tremor downtown. Looks like someone dropped something HUGE. \myemoji{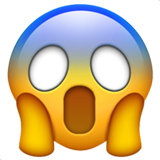} \#SanFrancisco; Location `San Francisco' not found in tweet\\
    1 & 1 & 0 & 1 & Generated tweet: Downtown San Francisco is crumbling after the quake! \myemoji{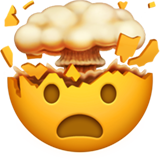} Massive structural damage reported, with officials saying damage is minimal. \#SanFrancisco \#Earthquake; Damage mismatch: expected `0', predicted `1'\\
    2 & 1 & 0 & 1 & Generated tweet: Downtown San Francisco is showing significant cracks and deformation. Emergency services are responding to the situation. \#SanFrancisco \#Earthquake; Damage mismatch: expected `0', predicted `1'\\
    3 & 1 & 0 & 0 & Generated tweet: Downtown San Francisco is cracking with damage. Authorities confirm structural damage, with significant visible cracks. \#SanFrancisco \#Earthquake; Damage level: 1; Damage mismatch: expected `0', predicted `1'; Too similar to accepted corpus (Self-BLEU=47.9 $>$ 40.0)\\
    \bottomrule
    \end{tabular}}
    \caption{Examples of the feedback text $f$, which includes corresponding compliance messages based on the compliance results for location correctness ($c_{loc}$), damage level correctness ($c_{dmg}$), and textual diversity ($c_{div}$). These synthetic tweets were generated using the \texttt{gemma-3-1b-it} model with temperature $\tau=1.4$, given the target location $y_{loc}=San$ $Francisco$ and target damage level $y_{dmg}=0$.}
    \label{tab:feedback_examples}
\end{table}

\section{Different LLMs as tweet generator}\label{supp:llms}
The workflow was able to generate synthetic tweets that captured their prespecified target labels regardless of the LLM used as the tweet generator $g$, as shown in Supplementary Figure \ref{fig:s_llms} for the 2014 Napa, California earthquake event. For all LLMs, iterative feedback improved the percentage of synthetic tweets accepted to the synthetic tweet dataset. 

\begin{figure}[h!]
    \centering
    \includegraphics[width=0.9\linewidth]{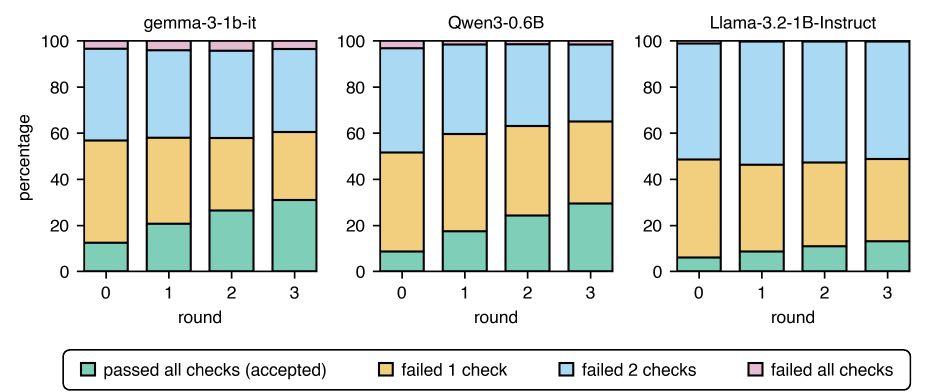}
    \vspace{0.4cm}
    \caption{Distribution of synthetic tweets by the number of compliance checks passed when using different LLMs (Google's \texttt{gemma-3-1b-it} model, Qwen's \texttt{Qwen-3-0.6B} model, and Meta's \texttt{Llama-3.2-1B-Instruct} model) as the tweet generator $g$. Results shown are based on the 2014 Napa, California earthquake event.}
    \label{fig:s_llms}
\end{figure}

\vspace{-0.5cm}
\section{Varying the LLM's temperature}\label{supp:temp}
Across all earthquake events, higher temperature values resulted in more synthetic tweets being accepted at each feedback round and overall, as shown in Supplementary Figure \ref{fig:s_temp}. Varying the LLM's temperature $\tau$ had a small effect on the checks for location and damage level correctness, but largely affected the textual diversity check.

\begin{figure}[h!]
    \centering
    \includegraphics[width=\linewidth]{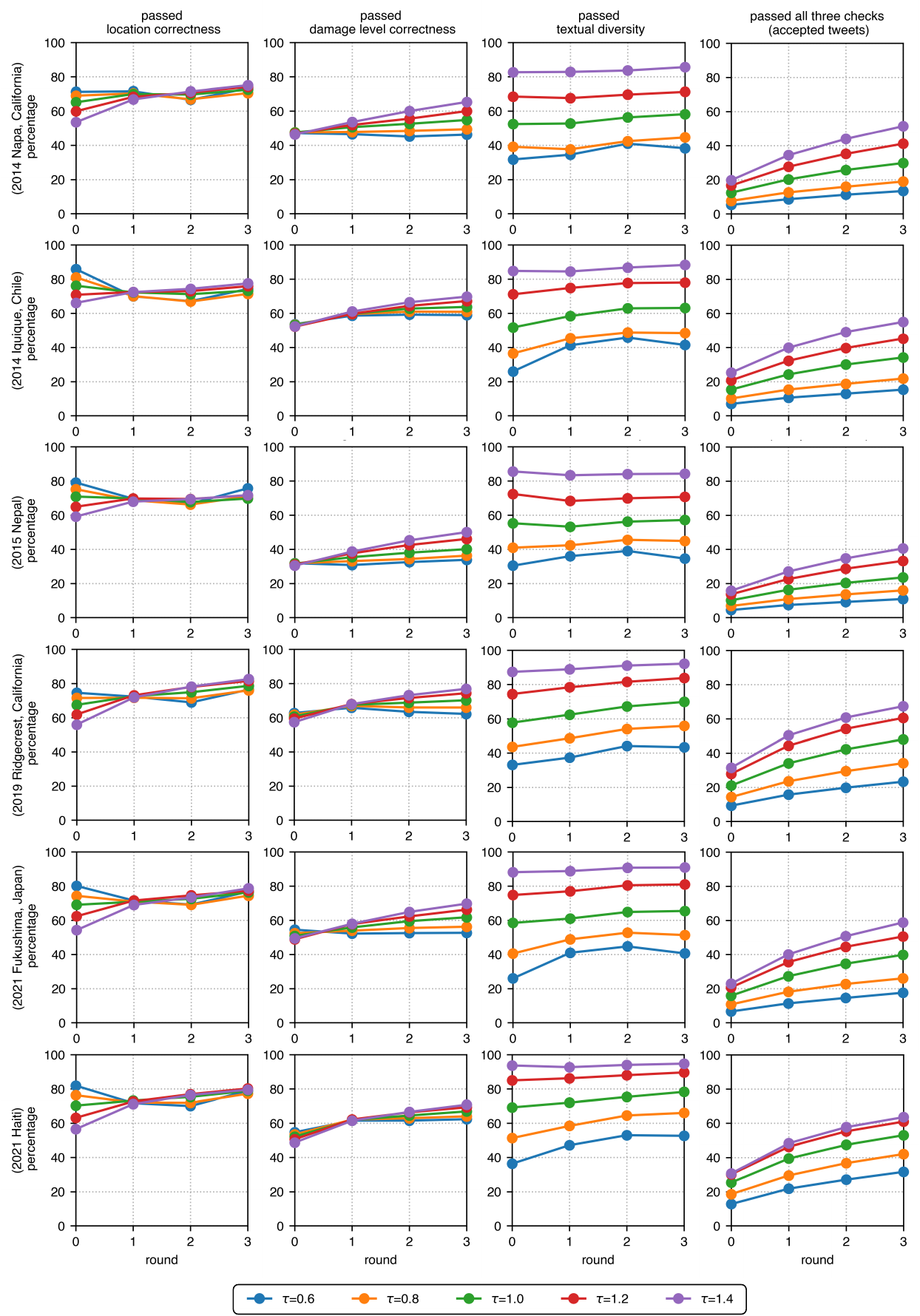}
    \caption{Percentage of tweets that passed the compliance checks when using different temperature $\tau$ values across the six earthquake events.}
    \label{fig:s_temp}
\end{figure}
\end{document}